\documentclass[conference]{IEEEtran}
\ifCLASSINFOpdf
\else
\fi

\usepackage{amssymb}
\usepackage{graphicx}
\usepackage{algorithm}
\usepackage{algorithmic}

\usepackage{url}
\usepackage{amssymb}
\usepackage{amsmath,bm}
\usepackage{color}
\begin{document}
%
\title{The Advantage of Evidential Attributes in Social Networks}



\author{\IEEEauthorblockN{Salma Ben Dhaou$^\text{a}$,
Kuang Zhou$^\text{c}$, 
Mouloud Kharoune$^\text{b}$, 
Arnaud Martin$^\text{b}$, and
Boutheina Ben Yaghlane$^\text{a}$
}
\IEEEauthorblockA{a. LARODEC, Higher Institute of Management, 41 Rue de la Liberte Cite Bouchoucha, 2000 Tunis, Tunisia}
\IEEEauthorblockA{b. DRUID, IRISA, University of Rennes 1, Rue E. Branly, 22300 Lannion, France}
\IEEEauthorblockA{c. Northwestern Polytechnical University, Xi'an, Shaanxi 710072, PR China.}
}


%


\maketitle

\begin{abstract}
Nowadays, there are many approaches designed for the task of detecting communities in social networks. Among them, some methods only consider  the topological graph structure, while others take use of both the graph structure and the node attributes. In real-world networks, there are many uncertain and noisy attributes in the graph. In this paper, we will present how we detect communities in graphs with uncertain attributes in the first step. The numerical, probabilistic as well as evidential attributes are generated according to the graph structure. In the second step,  some noise will be  added to the attributes. We perform experiments on graphs with different types of attributes and compare the detection results in terms of  the Normalized Mutual Information (NMI) values. The experimental results show that the clustering with evidential attributes gives better results comparing to those with probabilistic and numerical attributes. This illustrates the advantages of evidential attributes.
\end{abstract}

\begin{IEEEkeywords}
Belief function theory, uncertain attributes, community detection.
\end{IEEEkeywords}

%
\IEEEpeerreviewmaketitle

\section{Introduction}
Recently, social network analysis has become an important research topic. In fact, social network analysis can be defined as a distinct research perspective within the social and behavioral sciences. It is also based on the assumption of the importance of relationships among interacting units.

In social network analysis \cite{science-mono1, science-mono2}, the observed attributes of social actors are understood in terms of patterns or structures of ties among the units. These ties may be any existing relationship between units; for example friendship, material transactions, etc.

In this context, many studies use the concept of social network analysis to classify, for example the users's opinions~\cite{science-journal5} or to determine the true nature of a received message~\cite{proceeding7}.

However, using the techniques of the social network analysis in a real application can be a difficult task. Indeed, we will find ourselves in front of imprecise and uncertain information. This is the case of data collected through automated sensors for example \cite{jour8}.
For these reasons, it will be more interesting to use an attributed graph which is composed of weighted vertices and edges.
Although the majority of existing works in the literature focused on the study of weighted networks where the weights take integer values, recently, there has been studies of capturing nodes attributes via evidential models as presented in \cite{Jour21}.

However, as we are manipulating social data, there is always a probability to get errors in the observations or missing data.
In fact, the attributes can be constructed using statistical methods or maybe we are not totally sure about the type of the attribute. Indeed, for any node, we can have a vector of values composing its attribute. Hence, it will be interesting to use uncertain attributes in social networks.

In the same context, many studies focus on modeling the uncertain social network. In fact, they represent an uncertain network by weighting the nodes or links with values in $[0,1]$ to model uncertainties. Then, it will be more easier to monitor the behavior of the social network \cite{science-journal1,proceeding7}.

Nowadays, we can no longer talk about social networks without stating the concept of community detection.
Indeed, in all social networks, there is a group of individuals who are closely related to each other more than to others. This may be due to shared interests, practice, apprenticeship or preferences regarding particular topics.

According to Santo Fortunato~\cite{jour2}, communities, also called clusters or modules, represent groups of vertices which probably share common properties and/or play similar roles within the graph.
He argues also that the word community itself refers to a social context. In fact, people naturally tend to form groups, within their work environment, family or friends.

The community detection task becomes important since it allows us to classify the nodes according to their structural position and/or their
attributes. Indeed, the clusters obtained by the community detection algorithm contain similar objects.

The aim of this paper is to show how, from clustering uncertain attributes of the nodes, we can detect the existing communities in the graph. We will also show that, after adding some noisy uncertain attributes, the evidential generation of the attributes gives the best NMI values comparing to the numerical and the probabilistic versions.

This paper is structured as follows. In section~\ref{background}, we remind some basic concepts of the theory of belief functions and some community detection methods. Section~\ref{Process} will be dedicated to our contribution. Finally, section~\ref{Experimentations} will be devoted to the experimentations and section~\ref{conclusions} will conclude the paper.





\section{Background}
\label{background}
In this section, some basis of the belief functions theory will be recalled first.
Then, we will present a definition of an attributed graph. Finally, we will compare some community detection methods.


\subsection{Belief Functions Theory}

The belief functions theory allows explicitly to consider the uncertainty of knowledge using mathematical tools \cite{book,jour5}.
It is a useful and effective way in many fields of applications such as classification, decision making, representation of uncertain  and not accurate information, etc.

In fact, it is a suitable theory for the representation and management of imperfect knowledge. It allows to handle the
uncertainty and imprecision of the data sets, to fuse evidence and make decisions.

The principle of the theory of belief functions consists on the manipulation of functions defined on subsets rather than singletons as in probability theory. These functions are called mass functions and range from 0 to 1.

Let $\Omega$ be a finite and exhaustive set whose elements are mutually exclusive, $\Omega$ is called a frame of discernment. A mass function is a mapping
\[
   m:2^\Omega \rightarrow [0,1]
\]
such that
\begin{equation}
   \displaystyle  \sum_{X \in 2^\Omega} m(X)=1 \,\, \mbox{and} \,\, m(\emptyset)=0
\end{equation}
 The mass $m(X)$ expresses the amount of belief that is allocated to the subset $X$.
 We call $X$ a focal element if $m(X) > 0$.

A consonant mass function is a mass function which focal elements are nested $A_1 \subset A_2 \subset \dots \subset \Omega$.

\subsection{Attributed Graphs}
According to \cite{jour12}, an attributed graph $G_a=(V_a,E_a)$ can be defined as a set of attributed vertices $V_a=\{v_1,\ldots,v_p,\ldots,v_q,\ldots,v_n\}$ and a set of attributed edges $E_a=\{\ldots,e_{pq},\ldots\}$. The edge $e_{pq}$ connects vertices $v_p$ and $v_q$ with an attributed relation.


\subsection{Some Community Detection Methods with only Graphs Structures} 

In this section, we recall some methods which aim to find communities based on the network structure. In the literature, there are several studies such as the hierarchical clustering \cite{book4} which is a method based on the development of a measure of similarity between pairs of vertices using the network structure. The disadvantage of this technique consists on ignoring the number of communities that should be used to get the best division of the network.

The second type of methods is the algorithms based on edge removal. We present here two techniques:

 The algorithm of Girvan and Newman \cite{jour18} which is a divisive method, in which edges are progressively removed from a network. In addition, the edges to be removed are chosen by computing the betweenness scores. The final step consists on recomputing the betweenness scores following the removal of each edge. This algorithm does not provide any guide to how many communities a network should split into. it is also slow.

 The algorithm of Radicchi {\em et al.} \cite{jour16} is also based on iterative removal of edges but uses a different measure. It is based on counting short loops of edges in the network. This method has a principle disadvantage which consists on failing to find communities if the network containing few triangles in the first place.

 The last method is an approach which aim is to discover community structure based on the modularity Q \cite{jour17}. The quality is high for good community divisions and low for poor ones.




\subsection{Some Community Detection Methods with Graphs Structures and Attributes}
In this section, we introduce some community detection methods based on graph structure and attributes.

 The presented model in \cite{proceeding6} uses both informations. In fact, a unified neighborhood random walk distance measure allows to measure the closeness of vertex on an attribute augmented graph. Then, the authors uses a K-Medoids clustering method to partition the network into $k$ clusters.

 A second method presented in \cite{proceeding11} consists on a model dedicated to detect circles that combines network structure and user profile. The authors learns for each circle, its members and the circle-specific user profile similarity metric.

 A third method  presented in \cite{proceeding20} consists on dealing with the uncertainty that occurs in the attribute values within the belief function framework in the case of clustering.

 It is important to consider both structure information and attributes in order to detect the network communities. In fact, if one source of information is missing or noisy, the other can solve the problem.

\section{Proposed Process}
\label{Process}
\subsection{Graphs with Uncertain Attributes}
Generally, a social network is modeled by a graph \linebreak $G=(V,E)$ where $V$ is a set of vertices and $E$ a set of edges. However, such a representation does not take into account imperfections resulting from inaccurate and uncertain data.

Therefore, it will be interesting to combine the theory of graphs with the theories dealing with uncertainty like probability \cite{book3,proceeding9}, possibility or theory of belief functions \cite{proceeding7} in order to provide a general framework for an intuitive and clear graphical representation of real-world problems \cite{proceeding10}.

Therefore, an uncertain social network will be represented by the classic notation of a simple graph in addition of attributes defined in $[0,1]$ on the nodes and links.

In this paper, we want to show how we can detect communities for graphs with uncertain attributes. We precise that this is not a new method of community detection, but a way to consider these kind of data.

\subsection{Algorithm}

In the algorithm \ref{Algo1}, we propose a method of generating numerical, probabilistic and evidential attributes in order to find communities and show how different attributes make it possible to place each node in its true community.

In the first step, we give a numerical attribute to each node (a single value $x \in[0,1]$) which indicates the membership of that node to the community according to the number of communities. We consider the node's class $C_i$ among the set of $n$ possible classes according to the value of $x$:
$$x\in\left[ \frac{i-1}{n},\frac{i}{n}\right].$$

\textbf{First scenario:} We randomly generate the values ​​of the attributes for each node $v \in C_i$ of the graph. We consider three kind of attributes: numerical, probabilistic and evidential.
\begin{itemize}
 \item Numerical attribute: We generate a value $x$ in $\left[ \frac{i-1}{n},\frac{i}{n}\right]$ for $v$.
 \item Probabilistic attribute: We generate a value $x$ in $\left[ \frac{i-1}{n},\frac{i}{n}\right]$ corresponding to the probability $p(v \in C_i)$. For the $n-1$ other probabilities, we generate $n-1$ values in $[0, 1-x]$ that we associate randomly to the other classes. In order to normalize the probability we divide by the sum of the generated values. This process generates $n$ values $x_i$.
 \item Evidential attribute: We generate consonant mass function. First, we generate a value $x$ in $\left[ \frac{i-1}{n},\frac{i}{n}\right]$ corresponding to the probability $m(C_i)$. Then the mass of the $2^{n-1}$ other focal elements containing $C_i$ are generated in $[0, 1-x]$ and randomly associated to the focal elements. At last, we normalize the mass function as in the probabilistic case. This process generates $1+2^{n-1}$ values $x_i$.
\end{itemize}

\textbf{Second scenario:} In order to avoid the arbitrary level of value on the real class, we affect the highest value to the real class.
\begin{itemize}
 \item Numerical attribute: In that case, we have only one value, so this second scenario cannot concern the numerical attributes.
 \item Probabilistic attribute: We search the maximum of the $n$ values $x_i$, and we swap the values.
 \item Evidential attribute: We search the maximum of the $1+2^{n-1}$ values $x_i$, and we swap the values.
\end{itemize}

After the generation of the attributes of each node, the community detection is made by the K-Medoids algorithm which is robust in the presence of noise. Moreover, this algorithm is interesting and effective in the case of small data. In the case of evidential attributes, we use the distance of Jousselme~\cite{jour20} between the attributes.

After that, we compare the obtained clusters with the real clusters. In order to measure the clustering quality in each cluster, we use the Normalized Mutual Information (NMI), a measure that allows a compromise between the number of clusters and their quality \cite{jour19}. The NMI is given by:
\begin{eqnarray}
NMI(A,B) & = & H(A)+H(B)/H(A,B)
\end{eqnarray}
with
\begin{eqnarray}
H(A) & = & - \sum_a P_A(a) \log P_A(a)\\
H(A,B) & = & - \sum_{a,b} P_{A,B}(a,b) \log P_{A,B} (a,b)
\end{eqnarray}

In a second step, in order to evaluate the robustness of the proposed approach, we select randomly few nodes of the graph and modify their class. Then, we compute again the NMI and compute the Interval of Confidence.

Algorithm \ref{Algo1} shows the outline of the process followed for evidential attributes in the second scenario.

\begin{algorithm}
\caption{Evidential Attributed Network - Adding Noisy Attributes}
\begin{algorithmic}
\REQUIRE G: Network, \\
 N: Number of vertices, \\
 K: Number of clusters, \\
 Ci: Elements of each cluster i\\
\ENSURE $nmiAttr$: Similarities between evidential attributes, $IC$: Interval of Confidence
\STATE // First Scenario: Random Generation
\FORALL{$v \in Ci$}
\STATE EvidentialLabels($C_i$)
\STATE // a function that generates randomly mass functions according to some conditions for each node belonging to $C_i$.
\STATE //Second Scenario
\STATE Sort(EvidentialLabels)
\STATE // Put the highest generated value on the attribute ``Ci" according to which community, the node belongs and the rest on the subsets containing ``Ci".
\ENDFOR
\STATE Use the K-medoids algorithm to cluster the nodes.
\STATE Compute the NMI (Normalized Mutual Information).
\STATE Compute the confidence interval.
\STATE // Adding Noisy Attributes in both scenarios.
\STATE Select randomly n nodes of the network and modify their attributes.
\STATE Use the K-medoids algorithm to cluster the nodes.
\STATE Compute the NMI.
\end{algorithmic}
\label{Algo1}
\end{algorithm}

%
%


\subsection{General Example}
Let G be a network with 2 communities:
\begin{itemize}
\item $C_1=\{1,2,3\}$
\item $C_2=\{4,5,6,7\}$
\end{itemize}
\subsubsection{Generation}
\quad

 \textbf{First Scenario}:
We start first by generating 3 types of attributes:
\begin{itemize}
\item Numerical Attributes: for the nodes of $C_1$, we set the value of $x$ be a real number in $[0,0.5]$ and for the nodes of $C_2$, the attributes will be set in $[0.5,1]$. For example, a node $v$ in $C_1$ can have an attribute value equal to $0.2$.
\item Probabilistic Attributes: for the nodes of $C_1$, we consider two values, $x$ which is in $[0,0.5]$ and the second one is $1-x$. For the elements of $C_2$ we do the same except that we choose the first value $x$ in $[0.5,1]$. For example, a node $v$ in $C_2$ can have an attribute values equal to $(0.6,0.4)$.
\item Evidential Attributes: for the nodes of $C_1$, we consider a mass function with only two focal elements: $x$ on $C_1$, picked in $[0,0.5]$ and one on $\Omega$. The second value will be equal to $1-x$. For example, a node $v$ in $C_1$ can have as attributes $(0,0.4,0,0.6)$.
 For the elements of $C_2$, we follow the same process, except that we generate the first value in $[0.5,1]$.
\end{itemize}
After that we use the K-medoids algorithm to cluster the nodes according to the three types of attributes. Then, we compare the obtained clusters with the real ones according to the NMI values. And finally, we compute the intervals of confidence.

\textbf{Second Scenario:}
In this scenario, we only consider probabilistic and evidential attributes.
From the previous generation, we affect the highest generated value to the real class. Let's consider a node $v$ in $C_1$ in the case of the probabilistic attributes. Let's assume that it has initially a couple of values $(0.2,0.8)$. Hence, in this scenario, the node $v \in C_1$ will have a new values $(0.8,0.2)$. In the case of evidential attributes, the node $v$ 
will have new sorted values $(0,0.6,0,0.4)$.

\subsubsection{Noisy Attributes}
\quad

Once the generation is done, we will select randomly few nodes of the network and we will modify their attributes in both of scenarios.

\textbf{First Scenario}:
We consider the random generation and choose for example to modify the attributes of one node \linebreak $v \in C_1$. Initially this node has $(0.2,0.8)$ as probabilistic attributes. Now, we modify that by selecting randomly a first value on $C_1$ in the interval $[0.5,1]$ instead of $[0,0.5]$. Hence, the node $v$ will have a new attributes values, for example $(0.7,0.3)$. We do the same thing for the evidential attributes. After that, we use the K-medoids to cluster the nodes and compute the NMI value.

\textbf{Second Scenario}:
We consider the sorted attributes. The same process as for the random generation will be followed for both probabilistic and evidential attributes.

\section{Experimentations}
\label{Experimentations}
In this section we will perform some  experiments on real networks from the UCI data sets, such as  the Karate Club network, the Dolphins network and the Books about US Politics network.

The Zachary Karate Club is a well-known social network 
studied by Zachary \cite{IEEEhowto:karate}. The study was carried out over a period of three years from 1970 to 1972.

In this network, we find:
\begin{itemize}
\item 34 nodes that represent the members of  Karate Club.
\item 78 pairwise links between members who are interacted outside the club.
\end{itemize}

During the study a conflict arose between the administrator ``John A" and instructor ``Mr. Hi", which led to the split of the club into two. Half of the members formed a new club around Mr. Hi, members from the other part found a new instructor or gave up karate.

\begin{figure}[!t]
\centering
\includegraphics[width=2.5in]{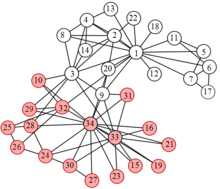}
\caption{The Karate Club Network.}
\label{karate_club}
\end{figure}

The Dolphins, animals social network introduced by Lusseau {\em et al.}~\cite{IEEEhowto:karate} is composed of 62 bottle-nose dolphins living in Doubtful Sound, New Zealand and social ties established by direct observations over a period of several years.

During the course of the study, the dolphins group split into two smaller subgroups following the departure of a key member of the population.

\begin{figure}[!t]
\centering
\includegraphics[width=3.2in]{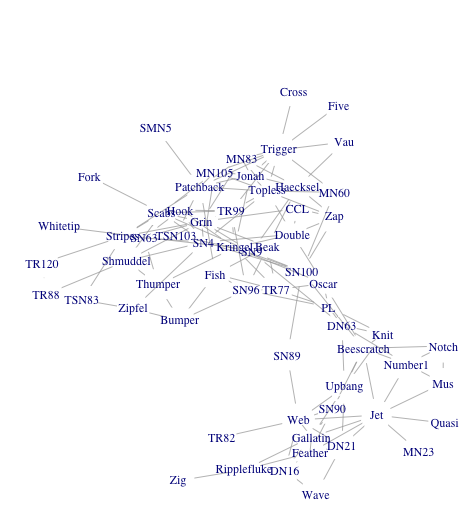}
\caption{Dolphins Network.}
\label{dolphins_network}
\end{figure}

The network of books \cite{IEEEhowto:karate} is composed of 105 nodes that represent books dealing with US politics sold by the on-line bookseller Amazon.com.
The edges represent frequent co-purchasing of books by the same buyers.

\begin{figure*}[!t]
\centering
\includegraphics[width=5.15in]{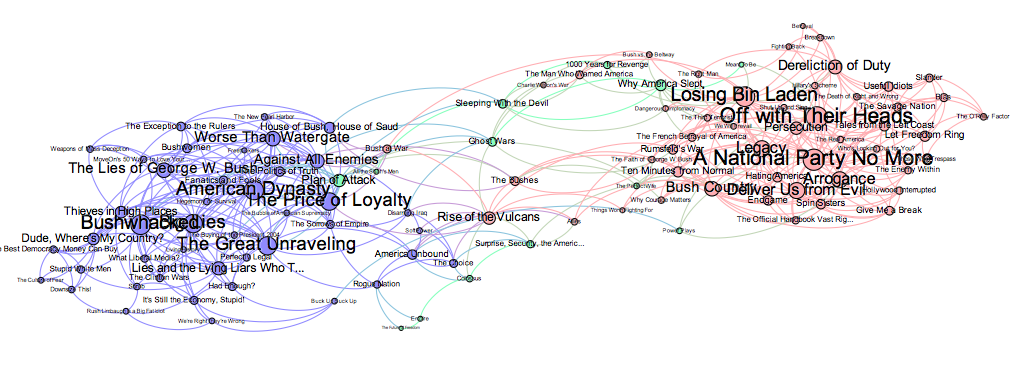}
\caption{Books about US Politics Network.}
\label{books_network}
\end{figure*}

\subsubsection{Process of Experimentations}
This experimentations allow us to show:
\begin{itemize}
\item How we can detect communities for graphs with uncertain attributes.
\item To what extent the uncertain attributes make it possible to find the communities after adding noisy data.
\end{itemize}
In this experimentation, we start first by generating attributes based on the structure of each network:
\begin{itemize}
\item Numerical Attributes: We remind that for this type of attribute, we generate a single value.
\begin{enumerate}
\item Karate Club: This network has 2 communities, so we give a single value of attribute to each node belonging to $C_1$ in the interval $[0,0.5]$ and a value in $[0.5,1]$ if the node belongs to $C_2$.
\item Dolphins Network: This network has also 2 communities. We choose the same intervals as for the Karate Club: if the node belongs to $C_1$, we generate an attribute in $[0,0.5]$ in $[0.5,1]$ if the node belongs to $C_2$.
\item Books about US Politics Network: This network has 3 communities: For the node belonging to $C_1$, we give an attribute in $[0,0.33]$. Each node belonging to $C_2$ has an attribute in $[0.33,0.66]$. Finally, for the nodes of $C_3$, they have an attribute in $[0.33,1]$.
\end{enumerate}
\item Probabilistic Attributes: For this type of attributes, we generate 2 or 3 values depending on the type of network.
\begin{enumerate}
\item Karate Club: For the nodes belonging to $C_1$, they have a first value picked randomly in the interval $[0,0.5]$ and the second value is deduced from that $(1-x)$. For the elements of $C_2$, the first values of attributes was picked randomly from the interval $[0.5,1]$ and the second one is deduced from that $(1-x)$.
\item Dolphins Network: Same things as for the karate club, the nodes of $C_1$ have a first value of attribute in $[0,0.5]$ and the second value is deduced from that. The nodes of $C_2$ have a first value in $[0.5,1]$ and the second one is deduced from the first one.
\item Books about US Politics Network: For the nodes of $C_1$, their first value of attributes will be picked in the interval $[0,0.33]$, the second and third values will be generated randomly from $[0,(1-x)]$. After that, we normalize by dividing the second and the third prob by the sum of the first, second and third probabilities.
 For the nodes of $C_2$, we followed the same process and for the elements of $C_1$, except that we picked the first values of the attributes in the interval $[0.33,0.66]$.
We note the same thing for the elements of $C_3$ except that we picked the first value in $[0.66,1]$.
\end{enumerate}
\item Evidential Attributes: For this type of attributes we generate 2 and 4 values, depending on the type of network.
\begin{enumerate}
\item Karate Club: This network has 2 communities so, $\Omega=\{C_1,C_2\}$ and $2^\Omega=\{\emptyset, C_1, C_2, C_1 \cup C_2\}$.
We choose to put 2 values on $C_1$ and $\Omega$ for the nodes belonging to $C_1$. For the rest of hypothesis, we put $0$. For the value of $C_1$, it  was picked in the interval $[0,0.5]$ and the second value on $\Omega$  was deduced from the first value. We remind that the sum should be equal to $1$. For the nodes of $C_2$, we put 2 values on $C_2$ and $\Omega$. The first value of $C_2$ is picked in $[0.5,1]$ and the second one is deduced of the first value.
\item Dolphins Network: We did the same thing with the nodes of this network as the Karate Club.
\item Books about US Politics: this network has 3 communities so,  $\Omega=\{C_1,C_2, C_3\}$ and
 $2^\Omega=\{\emptyset, C_1, C_2, C_1 \cup C_2, C_3, C_1 \cup C_3, C_2 \cup C_3, C_1 \cup C_2 \cup C_3\}$.
We choose to put 4 values on $C_1$, $C_1 \cup C_2$, $C_1 \cup C_3$ and $\Omega$ when the nodes belongs to $C_1$. For the rest of the hypothesis, we put the value $0$. The value of $C_1$ is picked from $[0,0.33]$ and the rest of values is deduced from the first one. We used the same principle as deducing the rest of probabilities presented previously, except that we generated 3 other probabilities instead of 2.
 For the second community, we did the same thing, except that we put values on $C_2$ and the subsets containing $C_2$. The value of $C_2$ is picked in $[0.33,0.66]$ and the rest of the values was deduced as explained before.
  For the third community, the values are generated on $C_3$, and each subset contain $C_3$. The value of $C_3$ is picked in $[0.66,1]$ and the rest of values is deduced as explained before.
\end{enumerate}
\end{itemize}
Once the attributes generated, we use the K-medoids algorithm to cluster the nodes according to their attributes. After that, we use the NMI method to compare the detected clusters with the real clusters of each network. Then, we compute the confidence interval.
These experimentations are repeated 100 times.
 In a second time, we sort the generated matrices by putting the highest values on $C_1$ and $C_2$ in the case of the Karate Club and Dolphins network and on $C_1$, $C_2$ and $C_3$ in the case of the Books about US Politics network. After that, we cluster again the nodes according to their new attributes and compute the NMI average.
 The second part of the experimentation consists on adding some noisy attributes by modifying the attributes of some nodes of $C_1$, $C_2$ and $C_3$. For each noisy attribute, we choose its value outside the interval set for its class. Then we cluster the nodes according to their attributes and compute the NMI and the interval of confidence. We perform this experimentation for the random and the sorted matrix of attributes.
 We precise that we use the sorted attributes matrices in the case of the probabilistic and the evidential generation.
 In the results below, we present the average of NMI computed for 100 executions of the experimentation and the interval of confidence for the numerical, probabilistic and evidential attributes.

\subsection{Comparison between the different versions of the labeled networks}
\subsubsection{Karate Club (First Scenario)}
In this section, we show the results of the NMI computation of the random generated attributes. We present below the results of the average values of NMI for 100 runs of random attributes generation.\\

\begin{tabular}{|c|c|c|}
\hline
 & NMI-Average &  Interval of Confidence \\
\hline
 Numerical & 0.776 & [0.596,0.955] \\
\hline
 Probabilistic & 0.778 & [0.59,0.967] \\
\hline
 Evidential & 1 & [1,1] \\
\hline
\end{tabular}
~\\~\\
The results show that the evidential generated attributes give better results than the probabilistic and the numerical ones. In fact, we obtained a value of the NMI average equal to 1 which means that the K-medoids is able to classify the nodes according to their evidential attributes in the right cluster even when the generation is random.
\\

\subsubsection{Dolphins (First Scenario)}

We present below the average values of NMI for 100 runs of random generated attributes in the Dolphins network.\\

\begin{tabular}{|c|c|c|}
\hline
 & NMI-Average & Interval of Confidence  \\
\hline
 Numerical & 0.782 & [0.587,0.976] \\
\hline
 Probabilistic & 0.765 & [0.554,0.976] \\
\hline
 Evidential & 1 & [1,1] \\
\hline
\end{tabular}
~\\~\\
We notice that the average evidential NMI is the highest value comparing to the probabilistic and the numerical ones. Same thing, the K-medoids is able to classify the nodes in their right cluster based on their evidential attributes.\\

\subsubsection{Books about US Politics: First Scenario}

In this section, we show the obtained results of the NMI average values in the case of 100 runs of random generated attributes.\\

\begin{tabular}{|c|c|c|}
\hline
 & NMI-Average & Interval of Confidence  \\
\hline
 Numerical & 0.699 & [0.551,0.848] \\
\hline
 Probabilistic & 0.758 & [0.668,0.848] \\
\hline
 Evidential & 1 & [1,1] \\
\hline
\end{tabular}
~\\~\\

The results show that the clustering based on the generated evidential attributes gives better results than the probabilistic and the numerical ones. In fact, the evidential NMI average is equal to one which means that all the nodes were classified in their right cluster.
\\

\subsubsection{Karate Club (Second Scenario)}

We executed the generation of the attributes several time and we sorted the matrix of attributes (We put the highest value on the attribute $C_1$ or $C_2$ depending on the belonging of the node to $C_1$ or $C_2$). We obtain the results of the average values of NMI for 100 executions below: \\

\begin{tabular}{|c|c|c|}
\hline
 & NMI-Average &  Interval of Confidence  \\
\hline
 Probabilistic & 0.7843 & [0.602,0.966] \\
\hline
 Evidential & 1 & [1,1] \\
\hline
\end{tabular}
~\\~\\
The results show that the evidential version gives an average NMI value equal to 1, which means that each node was detected in the right cluster. We notice that after sorting the probabilistic attributes, the K-medoids was not able to affect all the nodes in their right cluster.
\\

\subsubsection{Dolphins (Second Scenario)}

We proceed to sort the matrix of generated attributes and we compute the average values of NMI for 100 executions.\\

\begin{tabular}{|c|c|c|}
\hline
 & NMI-Average &  Interval of Confidence  \\
\hline
 Probabilistic & 0.79 & [0.597,0.983]\\
\hline
 Evidential & 1 & [1,1] \\
\hline
\end{tabular}
~\\~\\
We notice that the evidential version gives an average NMI value equal to $1$ comparing to the probabilistic and numerical versions. We also notice that the K-medoids was not able to classify the nodes in their right clusters based on their probabilistic attributes.
\\

\subsubsection{Books about US Politics (Second Scenario)}

We executed the generation of the attributes several times and we sorted the matrix of attributes (We put the highest value on the attribute $C_1$, $C_2$ or $C_3$ depending of the belonging of the node to $C_1$, $C_2$ or $C_3$). We obtain the results of the average values of NMI for 100 below:
\\

\begin{tabular}{|c|c|c|}
\hline
 & NMI-Average & Interval of Confidence \\
\hline
 Probabilistic & 0.895 & [0.828,0.962] \\
\hline
 Evidential & 1 & [1,1] \\
\hline
\end{tabular}
~\\~\\
The results show that the evidential version gives an average NMI value equal to 1 comparing to the probabilistic one which means that all the nodes were classified in their right clusters.

\subsection{Comparison between the different versions of the labeled networks after adding the noisy attributes} 

In this section, we present the obtained results after adding some noisy attributes. To do so, we choose randomly 1 to 9 nodes of the networks on which we add some noise. Hence, we modify their attributes values and we compute each time the NMI average values. This experimentation is repeated 100 times for each number of modified nodes, for cross-validation.



\begin{figure}[!t]
\centering
\includegraphics[width=3.5in]{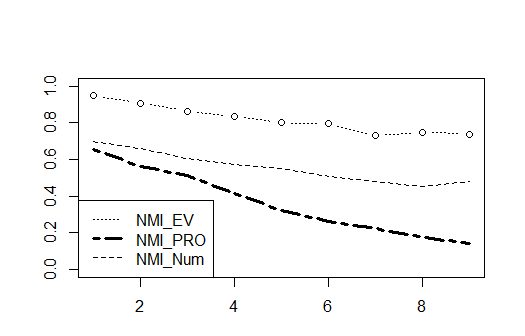}
\caption{Noisy Karate: First Scenario.}
\label{Noisy_karate_club}
\end{figure}




\begin{figure}[!t]
\centering
\includegraphics[width=3.5in]{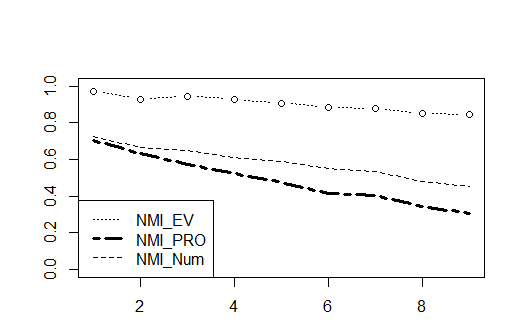}
\caption{Noisy Dolphins: First Scenario.}
\label{Noisy_Dolphins_Random}
\end{figure}

\begin{figure}[!t]
\centering
\includegraphics[width=3.5in]{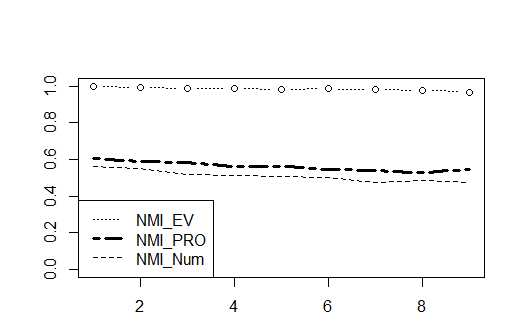}
\caption{Noisy Books: First Scenario.}
\label{Noisy_Books_Random}
\end{figure}

At first, we consider the first scenario and we present the results obtained on Karate Club dataset in figure \ref{Noisy_karate_club}, on Dolphins dataset in figure \ref{Noisy_Dolphins_Random} and on Books about US Politics dataset in figure \ref{Noisy_Books_Random}.

From the different curves, we deduce that the evidential attributes allow the K-medoids to cluster the nodes in their right clusters better than the numerical and the probabilistic attributes. In fact, we can notice that with the evidential attributes, almost all the nodes are classified in their right clusters even when the number of the noisy nodes is equal to $9$.
In addition, the intervals of confidence show that the evidential attributes are better than the probabilistic and numerical ones. For example, for $3$ noisy nodes, the interval of confidence in the case of the karate club network is equal to: $[0.408,0.809]$ for the numerical version, $[0.356,0.618]$ for the probabilistic version and $[0.711,0.967]$ for the evidential version.
 In the case of the Dolphins network, the interval of confidence is equal to: $[0.425,0.784]$ for the numerical version, $[0.467,0.908]$ for the probabilistic version and $[0.862,1]$ for the evidential version.
 Moreover, in the case of the Books about US Politics network, the interval of confidence is equal to: $[0.411,0.685]$ for the numerical version, $[0.533,0.646]$ for the probabilistic version and $[0.965,1]$ for the evidential version.


\begin{figure}[!t]
\centering
\includegraphics[width=3.5in]{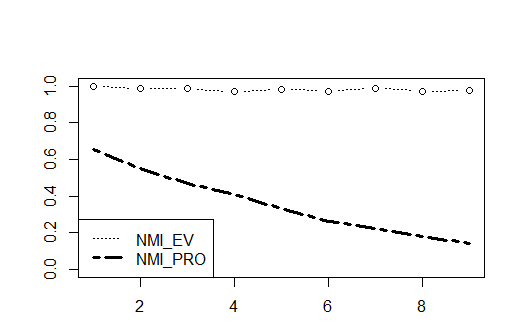}
\caption{Noisy karate: Second Scenario.}
\label{Noisy_karate_Sorted}
\end{figure}

\begin{figure}[!t]
\centering
\includegraphics[width=3.5in]{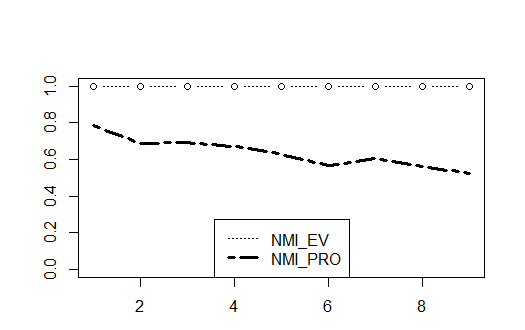}
\caption{Noisy Dolphins: Second Scenario.}
\label{Noisy_Dolphins_Sorted}
\end{figure}


\begin{figure}[!t]
\centering
\includegraphics[width=3.5in]{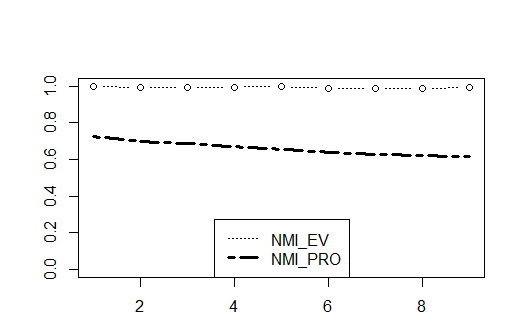}
\caption{Noisy Books: Second Scenario.}
\label{Noisy_Books_Sorted}
\end{figure}

Now, we consider the second scenario and we present the results obtained on Karate Club dataset in figure \ref{Noisy_karate_Sorted}, on Dolphins dataset in figure \ref{Noisy_Dolphins_Sorted} and on Books about US Politics dataset in figure \ref{Noisy_Books_Sorted}.

The results show that the clustering based on the evidential attributes gives better results than the probabilistic attributes. Indeed, the nodes with the evidential attributes are almost all classified in their right clusters. In addition, the intervals of confidence show that the evidential attributes are better than the probabilistic ones. For example, for $3$ noisy nodes, the interval of confidence in the case of the karate club network is equal to: $[0.467,0.793]$ for the probabilistic version and $[0.946,1]$ for the evidential version. In the case of the Dolphins network, the interval of confidence is equal to: $[0.515,0.791]$ for the probabilistic version and $[1,1]$ for the evidential version. And in the case of the Books about US Politics network, the interval of confidence is equal to: $[0.629,0.75]$ for the probabilistic version and $[0.963,1]$ for the evidential version.


\section{Conclusion}
\label{conclusions}
Throughout this paper, we reminded in the second section some basic concepts of the theory of belief functions and presented some communities detection methods. In the third section, we introduced our contribution consisting on communities detection based on the uncertain attributes of the vertices. In addition, we introduced some noisy attributes and compute the NMI values and the interval of confidence in order to show how uncertain attributes on nodes can be useful to find communities and show which type of attributes gives the best results.
We applied our algorithm on three real social networks: the Karate Club, the Dolphins network and the Books about US Politics.

Finally, in the experimentations section, we showed the results of each step used in our work.
In both scenarios, the evidential attributes generation gives a good results comparing to the probabilistic and the numerical attributes. In fact, in the random and sorted generation of attributes, when we cluster the nodes with evidential attributes, we obtained an average of NMI for 100 runs equal to one which means that  the nodes are affected to their real cluster.
In addition, when we introduced the noisy data, in the case of the evidential attributes, we obtained an average of NMI almost equal to 1. We can conclude that the theory of belief functions is a strong tool to model imprecise and uncertain attributes in the social networks.






%

\end{document}